\documentclass[10pt,twocolumn,letterpaper]{article}

\usepackage{cvpr}              

\usepackage{graphicx}
\usepackage{amsmath}
\usepackage{amssymb}
\usepackage{booktabs}
\usepackage{multirow}
\usepackage[accsupp]{axessibility}

\usepackage[breaklinks=true,bookmarks=false]{hyperref}

\usepackage[capitalize]{cleveref}
\crefname{section}{Sec.}{Secs.}
\Crefname{section}{Section}{Sections}
\Crefname{table}{Table}{Tables}
\crefname{table}{Tab.}{Tabs.}

\def\cvprPaperID{3105} 
\def\confName{CVPR}
\def\confYear{2023}

\begin{document}

\title{DAA: A Delta Age AdaIN operation for age estimation via binary code transformer}

\author{Ping Chen\\
Institution1\\
Institution1 address\\
{\tt\small firstauthor@i1.org}
\and
Xingpeng Zhang\\
School of Computer Science, SouthWest Petroleum University\\
Chengdu, China\\
{\tt\small secondauthor@i2.org}
}
%
%
\author{
Ping Chen$^{1}$, Xingpeng Zhang$^{2*}$, Ye Li$^1$, Ju Tao$^1$,  Bin Xiao$^2$,  Bing Wang$^2$, Zongjie Jiang$^1$\\ 
	$^{1}$Jiayu Intelligent Technology Co.,Ltd. (Affiliated With Great Wall Motor Company Limited)\quad\\
	$^{2}$School of Computer Science, SouthWest Petroleum University, Chengdu, China\quad \\
	{\tt\small \{redcping, yale.li.cn\}@gmail.com\quad xpzhang@swpu.edu.cn}\\
}

\maketitle
\begin{abstract}
   Naked eye recognition of age is usually based on comparison with the age of others. However, this idea is ignored by computer tasks because it is difficult to obtain representative contrast images of each age. Inspired by the transfer learning, we designed the Delta Age AdaIN (DAA) operation to obtain the feature difference with each age, which obtains the style map of each age through the learned values representing the mean and standard deviation. We let the input of transfer learning as the binary code of age natural number to obtain continuous age feature information. The learned two groups of values in Binary code mapping are corresponding to the mean and standard deviation of the comparison ages. In summary, our method consists of four parts: FaceEncoder, DAA operation, Binary code mapping, and AgeDecoder modules. After getting the delta age via AgeDecoder, we take the average value of all comparison ages and delta ages as the predicted age. Compared with state-of-the-art methods, our method achieves better performance with fewer parameters on multiple facial age datasets.
\end{abstract}

\section{Introduction}
\label{sec:intro}

Facial age estimation has been an active research topic in the computer version, for its important role in human-computer interaction\cite{Fragopanagos2005Emotion, Shu2018Personalized}, facial attribute analysis\cite{Angulu2018survey, Merillou2008survey}, market analysis\cite{Angulu2018survey}, and so on. After the rise of deep learning, many deep structures, such as VGG\cite{Simonyan2015VGG}, ResNet\cite{He2016res}, MobileNet\cite{Sandler2018mobile}, have been used as feature learning methods to solve the problem of facial age estimation\cite{Yang2018SSR, Zhang2019C3AE, Deng2021PML}.

In general, the methods for facial age estimation can be grouped into three categories: regression methods, classification methods, and ranking methods\cite{Pan2018Mean, Li2019BridgeNet}. The age regression methods consider labels as continuous numerical values\cite{Guo2011Simultaneous, Niu2016Ordinal}. Except for the universal regression, researchers also proposed hierarchical models\cite{Han2015Demographic} and the soft-margin mixture of regression\cite{Huang2017Soft} to handle the heterogeneous data. Facial age classification approaches usually regard different ages or age groups as independent category labels\cite{Guo2009Human}, which can be divided into single-label learning and label distribution learning methods\cite{Deng2021PML}. The single label learning\cite{Guo2009Human, Rothe2018Deep} treats each age independently, ignoring the fact that facial images of similar ages are very similar. Label distribution learning methods\cite{Geng2010Facial, Geng2013Deep, He2017Data, Shen2021Deep} learn a label distribution that represents the relative importance of each label when describing an instance. This method is to compare the distance or similarity between the distribution predicted by the model and the actual distribution\cite{Deng2021PML}. Nevertheless, acquiring distributional labels for thousands of face images itself is a non-trivial task. The ranking approaches treat the age value as rank-ordered data and use multiple binary classifiers to determine the rank of the age in a facial image\cite{Chang2011Ordinal, Chang2015A, Chen2017Using}. 

\begin{figure*}[t]
	\centering
	\includegraphics[width=0.95\linewidth]{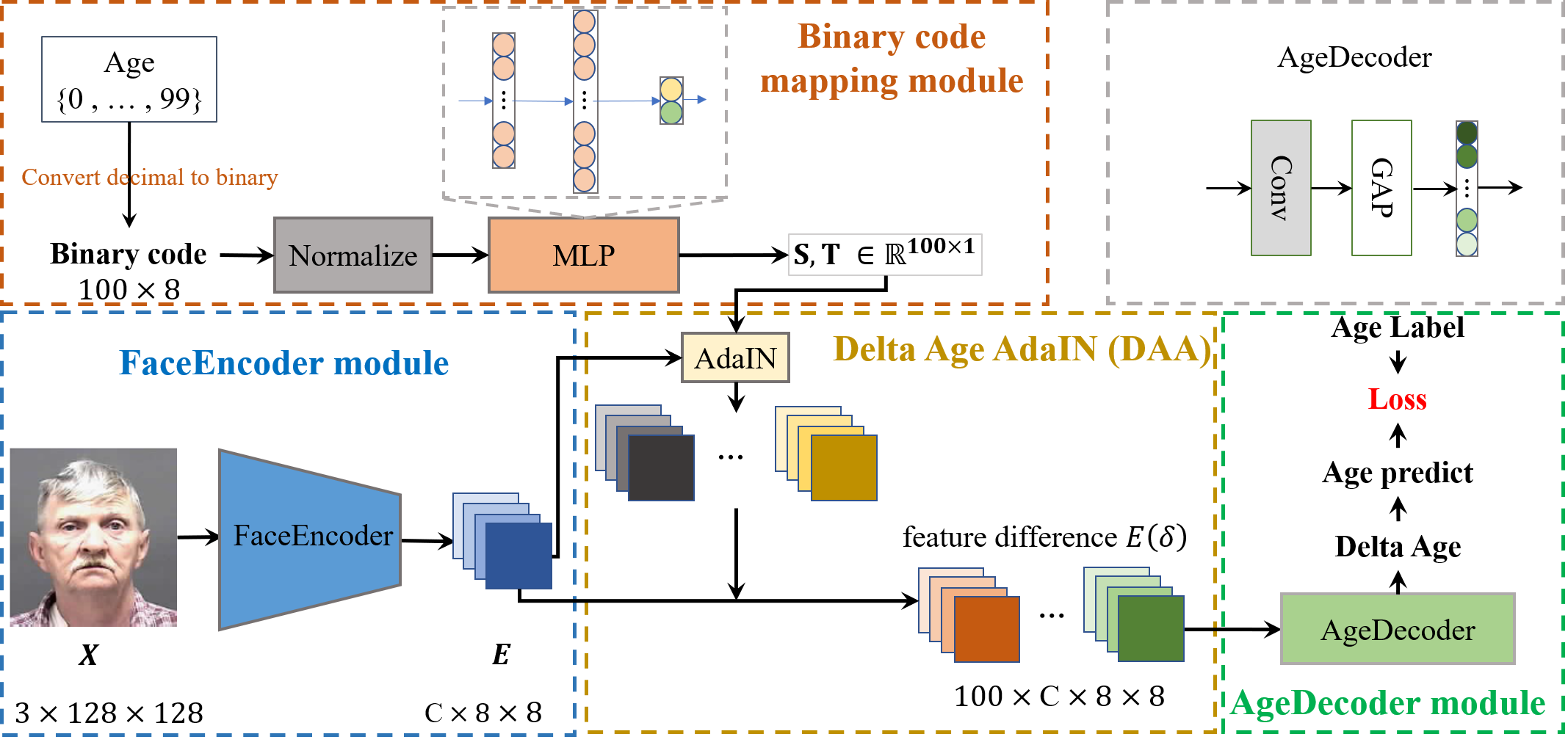}
	\caption{The overall structure of our network. The Network contains two inputs: the facial age image and the 8-bit binary code of ages. The MLP is a perception with three FC layers. FaceEncoder is a feature extraction block. Continuous feature differences between each age from 0 to 99 and the age of the input image are obtained by DAA transfers with the Binary code mapping module. And in the AgeDecoder module, more robust age estimation is done by the feature differences and their corresponding age label of binary codes.}
	\label{fig:onecol}
\end{figure*}

Although the above methods study the problem of facial age estimation from different emphases, they all belong to the perspective of computer vision, which can be summarized as feature extraction and modeling to predict age. This is different from the mechanism of the naked human eye recognizing age, which is obtained by comparing the current experience information with most humans. Because it is difficult to get representative age images of different races, computer tasks often ignore the idea of comparative learning. The style image can also be a contrast in style transfer learning.\cite{Karras2019style, Karras2020style2}. Inspired by this, we propose a Delta Age Adaptive Instance Normalization operation (DAA) to obtain representative results of each age through transfer learning. We want to transfer the current image into a style map of each comparative age. And then learn the feature difference between the current age and all the comparative ages. Finally, the predicted age is obtained based on the comparative age difference. Style images' mean and standard deviation are the keys to style transfer, and the random value cannot reflect the process of aging. We convert all ages into unique 8-bit binary codes and then learn comparative ages' mean and standard deviation vectors through the fully connected layer. 
The experiment results on four challenging age datasets demonstrate that our approach outperforms state-of-the-art methods.

The main contributions of this paper are as follows:
\begin{itemize}
   \setlength{\itemsep}{0pt}
	\setlength{\parsep}{0pt}
	\setlength{\parskip}{0pt}
	\setlength{\topsep}{0pt}
	\setlength{\partopsep}{0pt}
	\item  [$\bullet$] We designed the Delta Age AdaIN (DAA) operation based on the idea of human eye contrast learning. 
	\item [$\bullet$] To ensure that the delta age after transfer reflects continuity, we convert the natural number of ages into binary code. Finally, 100 delta ages feature maps will be generated for each content feature map.
	\item [$\bullet$] We designed a network based on age transfer learning to realize robust age estimation, achieving excellent performance on four datasets.
\end{itemize}

\section{Related works}
\label{sec:relate}

\subsection{Facial age estimation}
From the perspective of machine learning, facial age estimation can be regarded as two steps: feature extraction and modeling. The regression methods\cite{Guo2011Simultaneous, Niu2016Ordinal}, classification methods\cite{Guo2009Human, Rothe2018Deep}, and ranking methods\cite{Chang2011Ordinal, Chang2015A, Chen2017Using} for age estimation are paid more attention to put forward different research methods according to label information. Regression methods regard labels as continuous values, and classification regards labels as independent values. While the ranking approach treats labels as rank-order data. These methods gradually consider aging a slow and continuous process, which means processing label information is essential. Besides, some researchers learn label distribution to represent the relative importance of each label \cite{Geng2010Facial, Geng2013Deep, He2017Data, Shen2021Deep}, which can also be seen as a special facial age classification method. Label distribution is a hot research direction at present, but acquiring distributional labels for thousands of face images itself is a non-trivial task.

The main role of the deep learning method in age estimation is feature extraction. In Ranking-CNN \cite{Chen2017Using}, DEX \cite{Rothe2015DEX}, AP \cite{Zhang2017Quantifying}, DLDL \cite{Gao2017DLDL}, and other papers, the model usually adopts the deep structures such as AlexNet, VGG, and ResNet as the feature extraction module. In addition to feature extraction and modeling methods, some scholars also focus on the objective optimization function \cite{Pan2018Mean, Deng2021PML}. ML-loss \cite{Pan2018Mean} proposed mean-variance loss for robust age estimation via distribution learning. Deng et al. \cite{Deng2021PML} proposed progressive margin loss (PML) for long-tailed age classification, aiming to adaptively refine the age label pattern by enforcing a couple of margins.

The direct use of deep structure will cause huge model parameters, so many scholars try to compress the deep model structure for age estimation. Some lightweight structures are introduced into facial age estimation, such as OCRNN \cite{Niu2016Ordinal}, MRCNN \cite{Niu2016Ordinal}, MobileNet \cite{Sandler2018mobile}, and so on. Besides, Yang et al \cite{Yang2018SSR} proposed a compact soft stagewise regression network (SSR-Net), which reduced the parameters to $40Kb$. And Zhang et. al \cite{Zhang2019C3AE} propose an extremely compact yet efficient cascade context-based age estimation model(C3AE).

\subsection{Style transfer and adaptive instance normalization}
Style transfer is a fascinating work; the ideas contained in it are worth thinking about deeply. In 2016, Gatys et al. \cite{Gatys2016Image} realized style transfer by calculating two images' content and style distance. Ulyanov et al. \cite{Ulyanov2016Texture} proposed a trained generator, which applies batch normalization (BN). This research found that replacing BN with instance normalization (IN) can significantly improve the convergence speed \cite{Ulyanov2017Improved}. Dumoulin et al. \cite{Dumoulin2017Learned} found that images with different styles can be generated by using different scales and displacements during IN operation, also known as conditional instance normalization (CIN). Huang et al. \cite{Huang2017AdaIN} proposed that the artistic style of the image is the cross-spatial statistical information of each feature channel of the feature graph, such as mean and variance. Style transfer can be realized by transferring each channel's mean and standard deviation. This operation is named adaptive instance normalization (AdaIN), which ensures transferring any styles to the feature maps.

StyleGAN \cite{Karras2019style} draws on the idea of AdaIN style transfer, removes the traditional input, and proposes a style-based generator, which takes a learnable constant as input. It mainly controls the visual features represented by each level by modifying the input of each level separately without affecting other levels. And StyleGAN2 \cite{Karras2020style2} adjusted the use of AdaIN to avoid water droplet artifacts effectively. With the advantage of 
GAN network, some scholars try to generate high-quality facial age images. LATS\cite{OrEl2020Lifespan} presented a method for synthesizing lifespan age transformations. RAGAN\cite{Makhmudkhujaev2021RAGAN} introduces a personalized self-guidance scheme that enables transforming the input face across various target age groups while preserving identity.

\section{Proposed Approach}
\label{sec:Approach}

Fig.\ref{fig:onecol} shows the overall architecture of our proposed approaches, containing four components: FaceEncoder module, DAA operation, Binary code mapping module, and AgeDecoder module. 

Our designed network needs two modal inputs. One is the facial age images $X$, fed into the FaceEncoder module. The other is the 8-bit binary code $z$ of age natural number, entered into the binary code mapping module aiming to learn a set of data reflecting each age characteristic.

\subsection{FaceEncoder Module}
The FaceEncoder module aims to do feature extraction via deep learning models.
Let the input image of our approach as $X \in \mathbb{R}^{3 \times H \times W}$, where $3, H, W$ denotes the channel, height, and width, respectively. After the feature extraction structure, the output is as follows. 
\begin{equation}\label{eq-1}
	E = f^{E}(X)
\end{equation}
where $E = [E_{0}, E_{1}, \cdots, E_{C-1}] \in \mathbb{R}^{C \times h \times w}$. Except for the ResNet18, we also apply the C3AE\cite{Zhang2019C3AE}, which is a famous lightweight network for age estimation.

In the next subsection, we will do a transfer operation. Followed by AdaIN and StyleGAN, we need to calculate the mean and standard deviation value of $E_{c}\in\mathbb{R}^{1\times h\times w}$, and $c=\{0,1,\cdots,C-1\}$.

\subsection{Delta Age AdaIN operation}
Delta Age AdaIN (DAA) operation is the most essential operation in our method. As mentioned earlier, we hope to estimate the age by comparing the current image with the most representative images of all ages. However, usual characteristic information of all ages is usually challenging to obtain. In style transfer learning \cite{Huang2017AdaIN, Karras2019style, Karras2020style2}, the mean and standard deviation are considered to be the most representative of the image style. Inspired by this, we hope that the representative information of each age can be obtained via the transfer learning. Then, the feature difference was obtained by comparing the features of input age with transferred age. This is our proposed DAA operation. 

Let $E(x)=\{E_{0}(x), \cdots, E_{C-1}(x)\}$ be the learned feature $E$ of age $x$, $\mu(x)=(\mu_0(x), \mu_1(x), \cdots, \mu_{C-1}(x))$ and $\sigma(x)=(\sigma_0(x), \sigma_1(x), \cdots, \sigma_{C-1}(x))$ are the mean and standard deviation of feature $E(x)$, calculated as follows.
\begin{align*}
	\mu_c(x) & = \frac{1}{h\times w}\sum_{i=0}^{h-1}\sum_{j=0}^{w-1}E_{c}(x) \nonumber\\
	\sigma_{c}(x) & = \sqrt{\frac{1}{h\times w}\sum_{i=0}^{h-1}\sum_{j=0}^{w-1}(E_{c}(x)-\mu_c(x))^{2}+\epsilon}
\end{align*}

\begin{figure*}[t]
	\centering
	\includegraphics[width=0.95\linewidth]{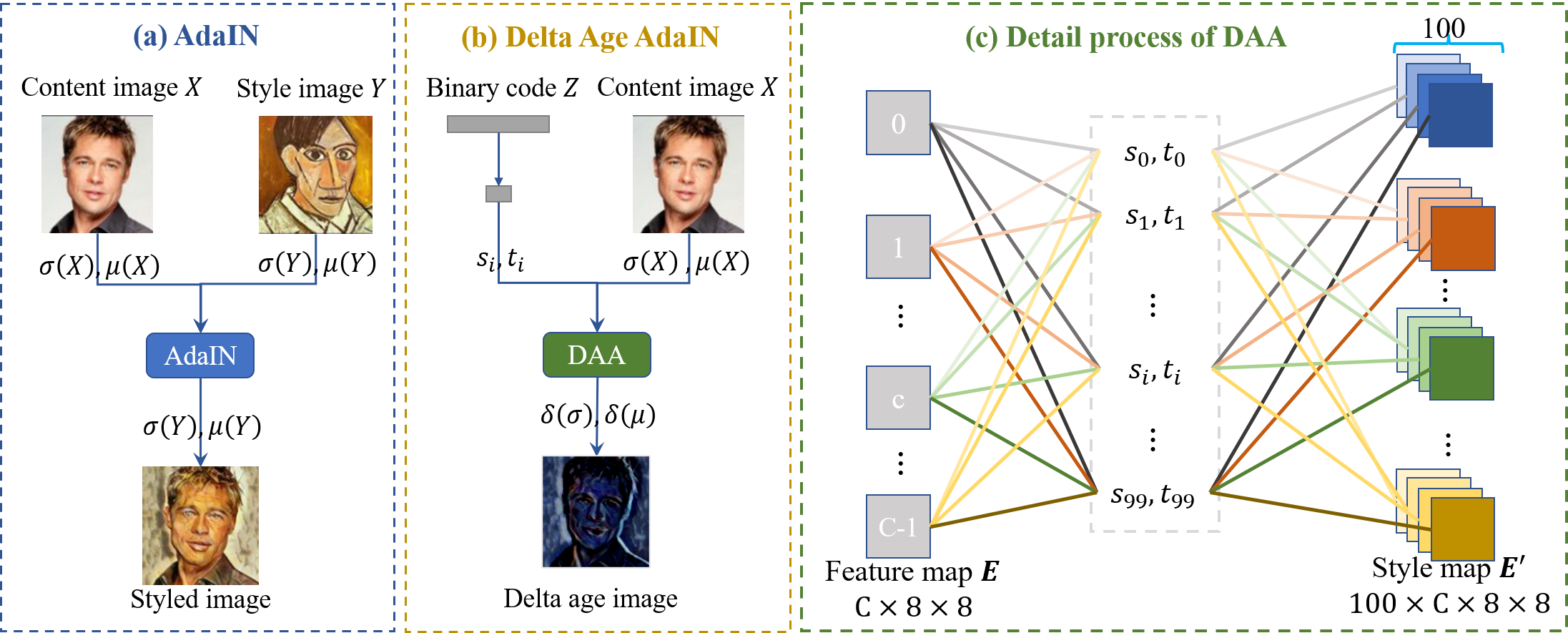}
	\caption{The DAA operation. (a) AdaIN operation; (b) Delta Age AdaIN (DAA); (c) Detail process of DAA.}
	\label{fig:DAA}
\end{figure*}

Inspired by literature \cite{Sveinn2018Generative,Yuval2021Only} based on style transfer to complete face aging, we use AdaIN \cite{Huang2017AdaIN} for age estimation.
The AdaIN has the following formula.
\begin{equation}\label{eq-2}
	AdaIN(E(x), E(y)) = \sigma(y)\frac{E(x)-\mu(x)}{\sigma(x)} + \mu(y)
\end{equation}
where $x$ is the age label of input $X$, and $y$ is the style age from 0 to 99, $E(x), E(y) \in\mathbb{R}^{C \times h \times w}$ can be seen as the feature of the content image and style image, respectively. This process is also shown in Fig. \ref{fig:DAA} (a).

We do not need to decode the style feature into the style image but combine it with the content feature to get the feature difference for estimating the age difference.
Let $AdaIN(E(x), E(y))$ denote the style feature of age x to y, and $AdaIN(E(x), E(x))$ denotes the content feature of age x. 
Following the facial aging process, we can get the feature difference of age difference by $AdaIN(E(x), E(y))-AdaIN(E(x), E(x))$. This process can be described with DAA operations as Eq.\eqref{eq_daa_src}.
\begin{equation}\label{eq_daa_src}
	\delta(x, y) = (\sigma(y)- \sigma(x))\frac{E(x)-\mu(x)}{\sigma(x)} + \mu(y)- \mu(x)   
\end{equation}
where $\delta(x, y)\in \mathbb{R}^{C \times h \times w}$ denote the feature difference between age $x$ and $y$. Eq. \eqref{eq_daa_src} is a single channel transfer strategy with each feature map having its own mean and standard deviation. This process is shown in Fig \ref{fig:DAA} (b). 

According to StyleGAN \cite{Karras2019style}, and SAM \cite{Yuval2021Only}, one channel of the deep feature can represent an attribute, so we consider all channel information simultaneously to obtain more representative comparative age transfer results. And the mean and standard deviation of the style feature of age $y$ can be recalculated as follows.
\begin{align}\label{eq-st}
	\mu_y & = \frac{1}{h\times w\times C}\sum_{i=0}^{h-1}\sum_{j=0}^{w-1}\sum_{c=0}^{C-1}E(y) \nonumber\\
	\sigma_y & = \sqrt{\frac{1}{h\times w\times C}\sum_{i=0}^{h-1}\sum_{j=0}^{w-1}\sum_{c=0}^{C-1}(E(y)-\mu_y)^{2}+\epsilon}
\end{align}
Then, the DAA operation can be rewritten as Eq.\eqref{eq-daa}
\begin{equation}\label{eq-daa}
	\delta_{c}(x, y) = (\sigma_y- \sigma_{c}(x))\frac{E_{c}(x)-\mu_{c}(x)}{\sigma_{c}(x)} + \mu_y- \mu_{c}(x)   
\end{equation}
where $\delta(x, y) = [\delta_{0}(x, y),\cdots, \delta_{C-1}(x, y)]$. Eq.\eqref{eq-daa} is a multi-channel transfer strategy with all feature maps sharing a mean and standard deviation. Let $f^{D}(\delta)$ be the nonlinear function that decodes feature difference into age difference, denoted as $y - x = f^{D}(\delta(x,y))$. 

Assuming that there are template images representing a history of 0 to 99 years, it is easy to obtain 100 sets of feature differences for each age image through DAA operation, shown as Fig. \ref{fig:DAA} (c). 
Then, a more robust age estimation method can be obtained through multiple feature differences and their corresponding template age labels.
\begin{align}\label{eq-age}
	x^{'} = \frac{1}{100}\sum_{y=0}^{99}(y - f^{D}(\delta(x,y)))
\end{align}
where $x^{'}$ is the predicted age for input $X$.

However, there are also enormous differences between images of the same age due to the influence of living environment, race, etc. Therefore, it is difficult to find an adaptive template for all ages. Consequently, we further design a binary code mapping module to replace the age templates to realize the robust age estimation by DAA operation.

\subsection{Binary code mapping}
In the DAA operation, the age is estimated through different age image templates, similar to the naked eye estimation. However, it is difficult to obtain representative contrast images.
Therefore, we hope to use random input latent code \cite{Karras2019style} to learn the mean ($\mu_y$) and standard deviation ($\sigma_y$) of style feature of age $y$. And the phenomenon that age is a gradual aging process also corresponds to the representative image of each age, i.e., it has the characteristics of continuity. Obviously, the random input cannot meet this requirement. 
Therefore, we replace the random latent code with an 8-bit binary code for all-natural age values based on the age range and the characteristics of binary coding. The uniqueness and continuity of binary coding are why we adopt this operation.
\begin{equation*}
	z_{y} = bin(y+1)
\end{equation*}
where $bin$ is a function that converts a decimal to binary. And $y\in\{0,1,\cdots,99\}$, $z=[z_{0}, \cdots, z_{99}]\in\mathbb{R}^{100\times 8}$, where $100$ corresponds to the age ranging from 0 to 99, and $8$ denotes the bit. Then, $z$ is normalized and got $Z_{0}$.

Similar to style transfer, we learn two 100-dimensional values, which correspond to the mean and standard deviation of the style features, respectively.
After three fully connected layers, the $Z_{0}$ is learned to two values.
\begin{equation*}
	Z_{i} = f^{B}_{i}(w_{i}*Z_{i-1}+b_{i})
\end{equation*}
where $i=1,2,3$ denotes the three FC layers, $f^{B}_{1}(\cdot)$ and $f^{B}_{2}(\cdot)$ apply the ReLU activation, $f^{B}_{3}(\cdot)$ applies identify function, $w_{i}$ and $b_{i}$ denotes the weight and bias of the $ith$ FC layer. For lightweight purposes, 
the number of nodes in three FC layers is $\{16, 32, 2\}$. We find that the reasoning speed of the binary module is very fast. Sometimes it only needs one training to get a better transfer value.

$Z_{3}$ can be expressed as two vectors $Z_{3} = \{S, T\}$, where $S=[s_{0}, s_{1}, \cdots, s_{99}]$ and $T=[t_{0}, t_{1}, \cdots, t_{99}]$.

After learning $S, T$ corresponding to each style age, the DAA operation Eq. \eqref{eq-daa} can be rewritten as Eq. \eqref{eq-4}
\begin{equation}\label{eq-4}
	\delta_{c}(x,y) = (s_{y} - \sigma_{c}(x))\frac{E_{c}(x)-\mu_{c}(x)}{\sigma_{c}(x)}+t_{y}-\mu_{c}(x)
\end{equation}
And $E(\delta) = [\delta(x,0), \cdots,\delta(x,99)]$ is the continuous feature differences between each age from 0 to 99
and age $x$, and $E(\delta)\in \mathbb{R}^{100 \times C \times h \times w}$.

It is worth noting that $Z_0$ is only used in training, and in the test phase, we only need to use $Z_3$ for DAA operations.

\subsection{AgeDecoder}
After the Binary code mapping module and DAA operation, we can get the continuous feature difference between ages 0 to 99. And we use AgeDecoder module to learn the nonlinear function $f^{D}(\delta))$ mentioned in Eq.\eqref{eq-age}. Our AgeDecoder module contains a nonlinear module and a linear regression module. Then, we get the age difference via the regression module.
\begin{align}
	\Delta & = f^{D}(E(\delta)) \nonumber\\
	& = Regression(GAP(Conv(E(\delta))))
\end{align}
where $\Delta=[\Delta_0,\cdots,\Delta_{99}]$, and  $\Delta_y\in\mathbb{R}^{1}$ is the delta age between age $x$ and the style age $y$ from 0 to 99. And $Regression$ is a fully-connected layer, $Conv$ denotes the convolution operation with a kernel size of 3, and its output channel is 64 in our experiments. $GAP$ is the global average pooling.

Then the Eq.\eqref{eq-age} can be rewritten as Eq.\eqref{eq_age_final}.
\begin{align}\label{eq_age_final}
	x^{'} = \frac{1}{100}\sum_{y=0}^{99}(y - \Delta_y)
\end{align}

The regression loss can be written as Eq.\eqref{eq_loss}.
\begin{equation}\label{eq_loss}
	Loss =SL1(x-x^{'}) 
\end{equation}
where $SL1(\cdot)$ denotes the smooth l1 loss.

During the training stage, we performed DAA operations for each age group to ensure continuity. In the test stage, we perform the DAA operation by selecting the age at equal intervals to speed up reasoning, which ensures continuity and robustness. For more details, see ablation experiments.

\section{Experiments}
\subsection{Datasets and Metrics}
\textbf{Datasets.} \emph{Morph} \cite{Ricanek2006MORPH} is the most popular dataset for facial age estimation, consisting of $55134$ face images from $13617$ subjects, and age ranges from $16$ to $77$ years. In our experiments, we follow the setting in C3AE \cite{Zhang2019C3AE}, where the dataset was randomly divided into the training part ($80\%$) and the testing part ($20\%$).

\emph{FG-Net} \cite{Panis2016Overview} contains $1002$ facial images from 82 subjects, where the age ranges from $0$ to $69$. In experiment, we adopt the setup of paper \cite{Guo2008Image,Pan2018Mean,Li2019BridgeNet,Deng2021PML}, which uses leave-one person-out (LOPO) cross-validation. Hence, the average performance over 82 splits is reported, which makes the time of each training longer, and the MAEs fluctuate greatly in each split.

\emph{IMDB-Wiki} \cite{Rothe2018Deep} consists of $523051$ facial images of celebrities, crawled from IMDB and Wikipedia, and age ranges from $0$ to $100$. Since there is much noise in this dataset, we selected about 300,000 images for training, where all non-face and severely occluded images were removed.

\emph{MegaAge-Asian}\cite{Zhang2017Quantifying} is newly facial age dataset consisting of 40,000 Asian faces with ages ranging from 0 to 70 years old. 
It helps to increase the diversity of human races and improve the generalization ability of the model. Followed by the setting of \cite{Zhang2017Quantifying, Yang2018SSR}, 3,945 images were reserved for testing, and the remains are treated as the training set. This dataset applies the cumulative accuracy (CA)\cite{Zhang2017Quantifying} as the evaluation metric.

\textbf{Metrics.} We take the Mean Absolute Error (MAE) in the experiments to calculate the discrepancy between the estimated age and the ground truth. Obviously, the lower the MAE value, the better performance it achieves. 

For the Mega-Age dataset, we also choose cumulative accuracy (CA) as the evaluation metric, which is defined as
\begin{equation*}
	CA(n) = \frac{K_{n}}{K}\times100
\end{equation*}
in which $K$ is the total number of testing images and $K_{n}$ represents the number of testing images whose absolute errors are smaller than $n$.

\subsection{Implementation Details}
The input images are cropped to $3\times 128 \times 128$. We randomly augmented all images in the training stage with horizontal flipping, scaling, rotation, and translation.

To reflect the universality and further highlight the performance of the proposed DAA method, we choose the classical network structure ResNet18 \cite{He2016resnet} with fewer parameters instead of ResNet34 and the lightweight network C3AE\cite{Zhang2019C3AE} for experiments on the Morph dataset. From the perspective of training time and video card resources, we choose ResNet18 with fewer parameters than ResNet34 as the backbone. And in the C3AE network, we do not use multi-scale but use the C3AE(plain) network. Limited by the length of the article, only the ResNet18 network is used for experiments on other datasets.
When applying ResNet18 and C3AE as the frame of the FaceEncoder module, we also do a few changes to satisfy the age image. The kernel size of the first convolution layer is changed to 3, and the first maximum pooling layer is discarded in ResNet18. Besides, the output $E$ of the FaceEncoder module using ResNet18 is the output of the last stage with the dimension of $512\times8\times8$. When using C3AE, the dimension of $E$ is $32\times8\times8$.

To have high performance and make a fair comparison, we also pretrained on the IMDB-Wiki dataset, which is similar to DEX\cite{Rothe2015DEX} and MV\cite{Pan2018Mean} methods.

For all experiments, we employed the Adam optimizer \cite{Kingma2015Adam}, where the weight decay and the momentum were set to $0.0005$ and $0.9$, respectively. Our DAA network is trained for 200 epochs with a batch size of 128. The initial learning rate was set to $0.001$ and changed according to cosine learning rate decay. We trained our model with PyTorch on an RTX 3080 GPU.

\subsection{Experiments Results and Analysis}
The DAA operation we designed can be transplanted to different feature extraction networks for age estimation. Our methods are expressed as \emph{Resnet18+DAA} and \emph{C3AE+DAA}. C3AE\cite{Zhang2019C3AE} is a lightweight network with a multi-scale image input, aiming to show that the designed DAA is also a lightweight operation. Following the design in C3AE\cite{Zhang2019C3AE}, we also give the "plain" configuration, applying single-scale input.

\textbf{Comparisons on Morph dataset.} As is shown in Table \ref{table_Morph}, 
the upper part of the table is the method with large parameters, while the lower part is the lightweight network. Compared with Ranking-CNN, AP, MV, and PML, our proposed method achieved the best MAEs value of $2.06$ with the pre-trained operation. At the same time, the parameters of our approach are significantly reduced compared with other methods. Compared with "VGG+Distillation" norm version, our DAA is lower but with much fewer parameters. 

\begin{table}
	\renewcommand{\arraystretch}{1.2}
	\caption{Comparision of MAEs and parameters on Morph dataset. ($^{*}$ indicates the result after pre-training, "norm" and "small" are two versions of "VGG+Distillation" method)}
\label{table_Morph}
\centering
\begin{tabular}{c|c|c}
	\hline
	Methods & MAE & Params \\
	\hline
	Ranking-CNN\cite{Chen2017Using} & 2.96 & 500M \\
	AP\cite{Zhang2017Quantifying} & 2.52 & 138M  \\
	MV\cite{Pan2018Mean} & 2.79/2.16$^{*}$ & 138M \\
	PML\cite{Deng2021PML} & 2.15$^{*}$ & 16M \\
    AVDL \cite{Wen2020AVDL} & 1.94$^{*}$ & 11M \\
    VGG+Distillation\cite{Zhao2021Distilling} (norm) & 1.95$^{*}$ & 69.5M \\
	\hline
	\textbf{ResNet18+DAA(ours)} & 2.25/{2.06$^{*}$} & {11M} \\
	\hline
	ORCNN\cite{Niu2016Ordinal} & 3.27 & 480K \\
	MRCNN\cite{Niu2016Ordinal} & 3.42 & 480K \\
	MobileNet\cite{Sandler2018mobile} & 6.50 & 226K  \\
	SSR\cite{Yang2018SSR} & 3.16 & 41K  \\
	C3AE\cite{Zhang2019C3AE} & 2.75 & 40K \\
	C3AE(plain)\cite{Zhang2019C3AE} & 3.13 & 40K \\
    VGG+Distillation\cite{Zhao2021Distilling} (small) & 2.73$^{*}$ & 0.11M \\
	\hline
	\textbf{C3AE(plain)+DAA(ours)} & \textbf{2.65} & 58K \\
	\hline
\end{tabular}
\end{table}

Our proposed method can also achieve excellent performance when applying C3AE (plain) \cite{Zhang2019C3AE} as the FaceEncoder module. Compared with ORCNN, MRCNN, and MobileNet, the MAEs value of our \emph{C3AE+DAA} reduced by more than $0.62$. Compared with only 40K lightweight network SSR and C3AE, our method has a similar amount of parameters, but the MAEs value is further reduced. Even compared with the C3AE network of multi-scale image input, the C3AE of single input using DAA operation is still reduced by $0.1$. Compared with "VGG+Distillation" small version, our DAA can achieve better performance with much fewer parameters.

\begin{table}
	\renewcommand{\arraystretch}{1.2}
	\caption{Comparison of MAEs and Parameters on FG-Net dataset. ($^{*}$ indicates the result after pre-training.)}
	\label{table_FG}
	\centering
	\begin{tabular}{c|c|c|c}
		\hline
		Methods & MAE & Params & Year\\
		\hline
		DEX\cite{Rothe2015DEX} & 4.63/3.09$^{*}$ & 138M & 2015\\
		DRFs\cite{Shen2021Deep} & 3.85 & - & 2021\\
		MV\cite{Pan2018Mean} & 4.10/2.68$^{*}$ & 138M & 2018 \\
		PML\cite{Deng2021PML} & 2.16$^{*}$ & 16M & 2021\\
		C3AE\cite{Zhang2019C3AE} & 2.95 & 40K & 2019\\
		\hline
		\textbf{ResNet18+DAA(ours)} & {2.19$^{*}$} & {11M} & -\\
		\hline
	\end{tabular}
\end{table}

\textbf{Comparisions on FG-Net.} As shown in Table \ref{table_FG}, 
we compared our model with the state-of-the-art models on FG-Net. 
Our DAA network achieves $2.19$ on the FG-Net dataset, which is only a little higher than the PML method in performance. It is better than the other pretrained model, such as MV-loss.

\begin{table*}
	\renewcommand{\arraystretch}{1.2}
	\caption{Comparision of CA on MegaAge-Asian dataset.}
	\label{table_MegaAge}
	\centering
	\begin{tabular}{c|c|c|c|c|c}
		\hline
		Methods & Pre-trained  & CA(3) & CA(5) & CA(7) & Year\\
		\hline
		Posterior\cite{Zhang2017Quantifying} & IMDB-WIKI & 62.08 & 80.43 & 90.42 & 2017 \\
		MobileNet\cite{Sandler2018mobile} & IMDB-WIKI & 44.0 & 60.6 & - & 2018 \\
		DenseNet\cite{Yang2018SSR}  & IMDB-WIKI & 51.7 & 69.4 & - & 2018\\
		SSR-Net\cite{Yang2018SSR}  & IMDB-WIKI  & 54.9 &74.1 & - & 2018\\
		LRN(ResNet10)\cite{Li2020LRN}& IMDB-WIKI & 62.86 & 81.47 & 91.34 &2020 \\
		LRN(ResNet18)\cite{Li2020LRN} & IMDB-WIKI & 64.45 & 82.95 & 91.98 & 2020 \\
		VGG+Distillation\cite{Zhao2021Distilling} & ImageNet, IMDB-WIKI, AFAD & 65.58 & 83.01 & 89.17 & 2021 \\
		\hline
		\textbf{ResNet18+DAA(ours)} & - & \textbf{68.82} & \textbf{84.89} & \textbf{92.70} & -\\
		\hline
	\end{tabular}
\end{table*}

Table \ref{table_FG} also shows the relationship between MAEs and parameters of different models. Compared with  MV models, our DAA model reduced the MAEs from $2.68$ to $2.19$ and decreased the parameters from 138M to 11M. Although the parameter of our method is higher than the C3AE model, the MAEs value decrease by $0.65$.

\textbf{Comparision on MegaAge-Asian dataset.} The experiment result of the Asian facial dataset is shown in Table \ref{table_MegaAge}. As seen from the table, the previous methods used pre-training operations on the IMDB-Wiki dataset. Compared with the pre-trained Posterior method, our approach increases $6.74$, $4.46$, and $2.28$ on each metric. On CA(3) and CA(5), our DAA model is $13.92$ and $10.75$ higher than SSR-Net. Compared with the distillation method which is pre-trained on ImageNet\cite{Deng2009ImageNet}, IMDB-WIKI, and AFAD\cite{Niu2016Ordinal}, our DAA method also achieves higher performance.

\textbf{Comparision on IMDB-clean dataset.} The experiment shown in Table \ref{table_IMDB} indicates that our approach can achieve a better MAEs value with much fewer parameters. Similar to the results on FG-Net, the parameter is higher than the C3AE model, and the MAEs value decrease by $5.17$.
\begin{table}
	\renewcommand{\arraystretch}{1.2}
	\caption{Comparison of MAEs on IMDB-clean dataset.}
	\label{table_IMDB}
	\centering
	\begin{tabular}{c|c|c|c}
		\hline
		Methods & MAE  & Params & Year\\
		\hline
		DLDL\cite{Gao2017DLDL} & 6.04 & 135M & 2017 \\
		DEX\cite{Rothe2015DEX} & 5.34 & 138M & 2015\\
		MV-loss\cite{Pan2018Mean} & 5.27$^{*}$ & 138M & 2018\\
		C3AE\cite{Zhang2019C3AE} & 6.75 & 40K  & 2018\\
		\hline
		\textbf{ResNet18+DAA(ours)} & \textbf{5.17} & 11M & - \\
		\hline
	\end{tabular}
\end{table}

\subsection{Ablation Study}
In the ablation experiment, we apply ResNet18 as the frame of the FaceEncoder module. Three experiments to illustrate the effectiveness of DAA operation, visualization of S, T value learned via Binary code mapping module, and comparison of reasoning time

\textbf{Experiment1: Effectiveness of DAA operation.}

We designed three baseline models to show the effectiveness of the DAA operation and binary code mapping. Baseline 1 is a network without the DAA operation. Baselines 2 and 3 adopt DAA operation using age template, corresponding to Eq. \eqref{eq_daa_src} and \eqref{eq-daa}, respectively. Because the age template is difficult to obtain, we adopted the average value of 10 random experiments. We randomly selected an image from each age as a template and continued to use these templates in subsequent training and testing.
The DAA operation with the binary transfer follows Eq. \eqref{eq-4}. 
\begin{table}
	\renewcommand{\arraystretch}{1.2}
	\caption{Effectiveness of DAA operation.}
	\label{table_templete}
	\centering
	\begin{tabular}{c|c|ccc}
		\hline
		\multirow{2}{*}{Methods} & Morph & \multicolumn{3}{c}{MegaAge-Asian}\\
		\cline{2-5}
		& MAE & CA(3) & CA(5) & CA(7) \\
		\hline
		w/o DAA & 2.72 & 63.12 & 80.15 & 90.39 \\
		\hline
		single channel  & 2.65 & 67.97 & 84.06 & 92.40 \\
		\hline
		multi-channel & 2.47 & 68.29 & 84.84 & 92.47 \\
		\hline
		Binary mapping & \textbf{2.25} & \textbf{68.82} & \textbf{84.89} & \textbf{92.70} \\
		\hline
	\end{tabular}
\end{table}

As shown in Table \ref{table_templete}, the performance of the three models using DAA has been improved. Single channel and multi-channel represent DAA operation using equations Eq. \eqref{eq_daa_src} and \eqref{eq-daa}, respectively. The experimental results show that the DAA operation using the mean of all channel information is more consistent with the style transfer of age. The DAA operation applying binary code mapping improved the Morph containing different races and the MegaAge dataset with fewer racial differences. This shows that the age templates that are difficult to obtain can be perfectly replaced by binary code mapping, thus reflecting the effectiveness of binary code mapping.

\begin{figure}[t]
	\centering
	\includegraphics[width=0.98\linewidth]{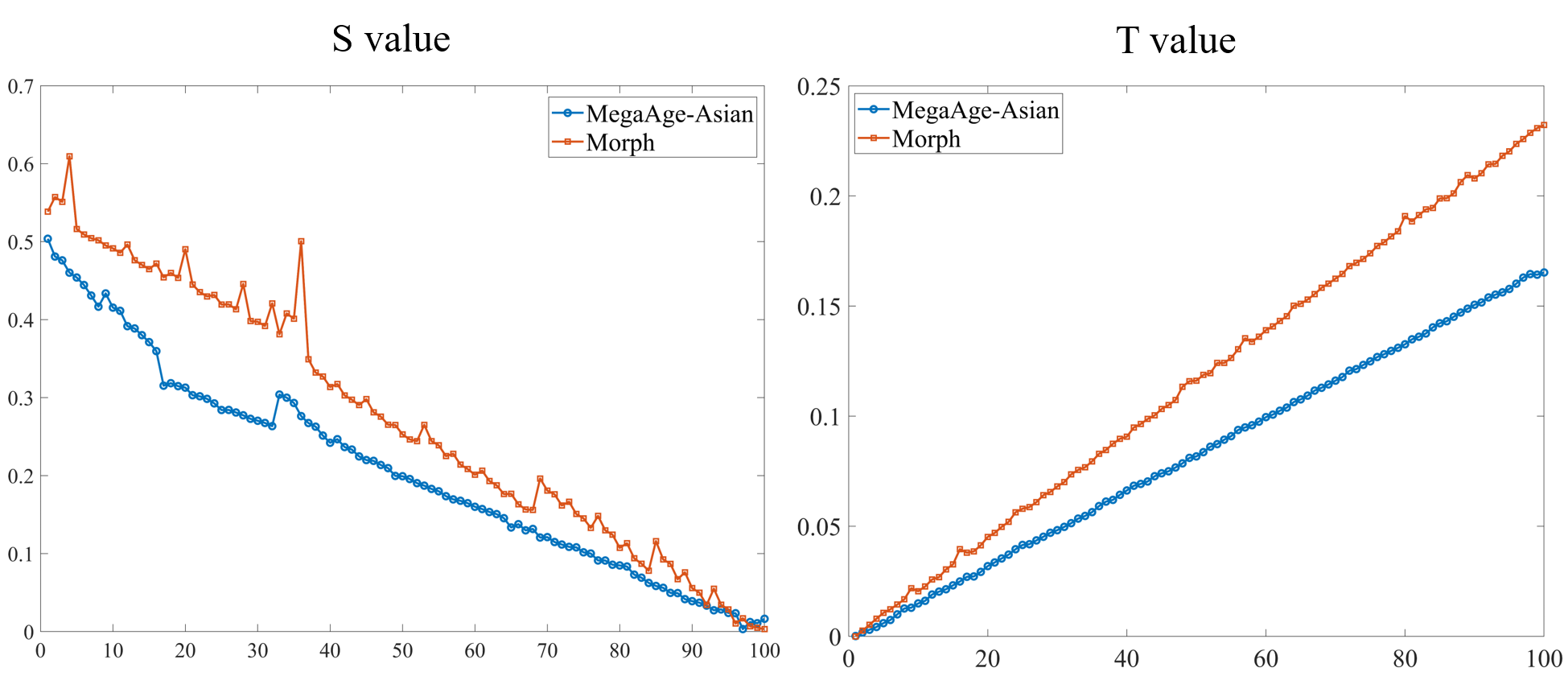}
	\caption{The learned S and T on MegaAge-Asian and Morph dataset. The horizontal axis is the age value of 0-99.}
	\label{fig2} 
\end{figure}

\textbf{Experiment 2: Learned two values in the Binary code mapping module.}

The effectiveness of binary code mapping has been demonstrated in Table \ref{table_templete}, and we further analyzed it by visualization, as shown in Fig. \ref{fig2}. We can draw interesting conclusions from the meaning of the normalized value $S$ and $T$. Each value on the $S$ and $T$ curve reflects the dataset's standard deviation and mean of the current age.
\begin{table}
	\renewcommand{\arraystretch}{1.2}
	\caption{The total reasoning time of DAA operation, Binary code mapping, and AgeDecoder on GPU/CPU.}
	\label{tab_time}
	\centering
	\begin{tabular}{c|c|c|c}
		\hline
		interval & sampling & GPU(ms) & CPU(ms) \\
		\hline
		1 & $S_{100}, T_{100}$ & 2.64 & 13.70  \\
		\hline
		2  & $S_{50}, T_{50}$ &2.51 & 10.75  \\
		\hline
		5  & $S_{20}, T_{20}$ & 2.75 & 10.35 \\
		\hline
		10  & $S_{10}, T_{10}$ & 2.50 & 10.17  \\
		\hline
		20  & $S_{5}, T_{5}$ & 2.59 & 8.59  \\
		\hline
		50 & $S_{2}, T_{2}$ & 2.58 & 8.58 \\
		\hline
	\end{tabular}
\end{table}

The $T$ curve shows that the mean value of the age image increases with age. This is consistent with the changing process of face images in reality. As age grows older, more and more wrinkles and spots will appear. This increases the complexity of the facial features, resulting in an increase in the average. Although the mean value increases, the difference between different images of the same age is getting smaller and smaller, which makes the standard deviation show a downward trend. The $S$ value we learned just fits this situation.
The two values in Morph dataset is higher than that in the MegaAge-Asian dataset. This is because Morph contains facial pictures of multiple races, and the amount is less.

\textbf{Experiment 3: Reasoning time analysis.}

According to the curve in Figure 3, we believe that the transfer process of DAA from a feature difference to an age difference is continuous and approximately linear. Then, we further verify
this phenomenon by selecting "S" and "T" in different parts,
whose results are shown in Table \ref{tab_time} and \ref{tab_ST}.  During the experiment, we use all ages during training and partial "S" and "T" values in the test.

To illustrate their continuity, we analyze the learned $S$ and $T$ with different intervals. Our method is plug and play, so the reasoning time here does not include FaceEncoder. Table \ref{tab_time} shows the running time of our approach with different intervals. 
The GPU is an RTX3080, and the CPU is Intel i5-8265U. In this table, "interval" represents the 100 $S$ and $T$ are grouped by the interval and then selected by sequence numbers. This process can be described as sampling, i.e. $S_{100/interval} = [s_{i* interval}]^{100/interval-1}_{i=0}$, $T_{100/interval} = [t_{i* interval}]^{100/interval-1}_{i=0}$. For example, when $interval =5$, $S_{20} = [s_{i* 5}]^{19}_{i=0}$.

As is exhibited in Table \ref{tab_time}, the running time of different intervals under different settings is almost the same. 

\textbf{Experiment 4: Continuity of learned two values.}
Table \ref{tab_time} and Table \ref{tab_ST} jointly show that the learned "S" and "T" have continuity, and sampling with intervals from 100 values can still have similar performance. The smaller the number of samples, the performance will be slightly reduced, but the reasoning time will be faster.

\begin{table}
	\renewcommand{\arraystretch}{1.2}
	\caption{The MAEs and CA of our approach under different numbers of learned $S, T$ on MegaAge-Asian dataset.}
	\label{tab_ST}
	\centering
	\begin{tabular}{c|c|c|c|c|c}
		\hline
		interval & sampling & MAE & CA(3) & CA(5) & CA(7) \\
		\hline
		1 & $S_{100}, T_{100}$ & 2.93 & 68.82 & 84.89 & 92.70 \\
		\hline
		2 & $S_{50}, T_{50}$ & 2.93 & 68.64 & 84.61 & 92.60 \\
		\hline
		5 &  $S_{20}, T_{20}$ & 2.94 & 68.75 & 84.84 & 92.52\\
		\hline
		10 & $S_{10}, T_{10}$ & 2.93 & 68.57 & 84.59 & 92.57\\
		\hline
		20 & $S_{5}, T_{5}$ & 2.93 & 68.62 & 84.56 & 92.57\\
		\hline
		50 & $S_{2}, T_{2}$ &2.95 & 68.62 & 84.59 & 92.37\\
		\hline
	\end{tabular}
\end{table}

\section{Conclusions}
In this work, we have proposed a Delta Age AdaIN operation (DAA) to obtain representative results of each age through transfer learning. The proposed DAA is a lightweight and efficient feature learning network. Our DAA will transfer each content map into $100$ delta age maps via learned $S$ and $T$ corresponding to the style of each age. We set the input of transfer learning to binary code form to obtain continuous image feature information. With the characteristic of the uniqueness and continuity of binary coding, we make the fused feature information continuous through DAA operation.
The designed module transfers the learned values in the binary code mapping module to feature maps learned by FaceEncoder module. 
Experiments on four datasets demonstrate the effectiveness of our approach. In future works, we will focus on the constraints in the training process to further improve the potential transfer effect. We will also discuss the probability distribution and corresponding interpretation of binary transfer.

\section*{Acknowledgement}
This work is supported by Natural Science starting project of SWPU (No.2022QHZ023) and Sichuan Scientific Innovation Fund (No. 2022JDRC0009). 

{\small
\bibliographystyle{ieee_fullname}
\bibliography{egbib}
}

\end{document}